\documentclass[letterpaper]{article} 

\usepackage{textcomp} 
\usepackage{aaai25}  
\usepackage{times}  
\usepackage{helvet}  
\usepackage{courier}  
\usepackage[hyphens]{url}  
\usepackage{graphicx} 
\usepackage{natbib}  
\usepackage{caption} 
\usepackage{algorithm}
\usepackage{algorithmic}
\usepackage{amsmath}
\usepackage{xcolor}
\usepackage{amssymb}
\usepackage{booktabs}
\usepackage{subcaption} 
\usepackage{graphicx} 
\usepackage{caption}
\usepackage{newfloat}
\usepackage{listings}
\DeclareCaptionStyle{ruled}{labelfont=normalfont,labelsep=colon,strut=off} 
\floatstyle{ruled}
\newfloat{listing}{tb}{lst}{}
\floatname{listing}{Listing}
\title{Reconsidering Feature Structure Information and Latent Space Alignment in Partial Multi-label Feature Selection}
\author{
    Hanlin Pan\textsuperscript{\rm 1,2},
Kunpeng Liu\textsuperscript{\rm 3},
Wanfu Gao\textsuperscript{\rm 1,2}\thanks{corresponding author}
}
\affiliations{
   \textsuperscript{\rm 1}College of Computer Science and Technology, Jilin University, China\\
   \textsuperscript{\rm 2}Key Laboratory of Symbolic Computation and Knowledge Engineering of Ministry of Education, Jilin University, China\\
  \textsuperscript{\rm 3}Department of Computer Science, Portland State University, Portland, OR 97201 USA\\
  panhl23@mails.jlu.edu.cn, kunpeng@pdx.edu, gaowf@jlu.edu.cn
}

\usepackage{bibentry}

\begin{document}

\maketitle

\begin{abstract}
The purpose of partial multi-label feature selection is to select the most representative feature subset, where the data comes from partial multi-label datasets that have label ambiguity issues. For label disambiguation, previous methods mainly focus on utilizing the information inside the labels and the relationship between the labels and features. However, the information existing in the feature space is rarely considered, especially in partial multi-label scenarios where the noises is considered to be concentrated in the label space while the feature information is correct. This paper proposes a method based on  latent space alignment, which uses the information mined in feature space to disambiguate in latent space through the structural consistency between labels and features. In addition, previous methods overestimate the consistency of features and labels in the latent space after convergence. We comprehensively consider the similarity of latent space projections to feature space and label space, and propose new feature selection term. This method also significantly improves the positive label identification ability of the selected features. Comprehensive experiments demonstrate the superiority of the proposed method.
\end{abstract}

%

\section{Introduction}

Partial Multi-Label Learning (PML) \cite{xie2018partial,yu2018feature} is a emerging framework within the realm of weakly supervised learning. In this paradigm, each instance is linked to a set of potential candidate labels, in which only part of these labels represent the true label. The characteristic of not requiring the label set to be completely correct makes this type of method more aligned with real-world scenarios and more robust in the face of noises in the dataset. It also diverges from traditional supervised learning methodologies, which precisely annotating label on each sample, thereby incurring substantial labeling cost. These two natures of PML renders it a viable solution for a multitude of practical applications, such as image recognition \cite{pham2022graph,zeng2013learning}, web mining \cite{chen2021predicting}, and ecoinformatics \cite{yilmaz2021multi,zhou2018weakly}.
\begin{figure}[h!]
    \centering
    \begin{minipage}{0.7\linewidth}
        \includegraphics[width=\textwidth]{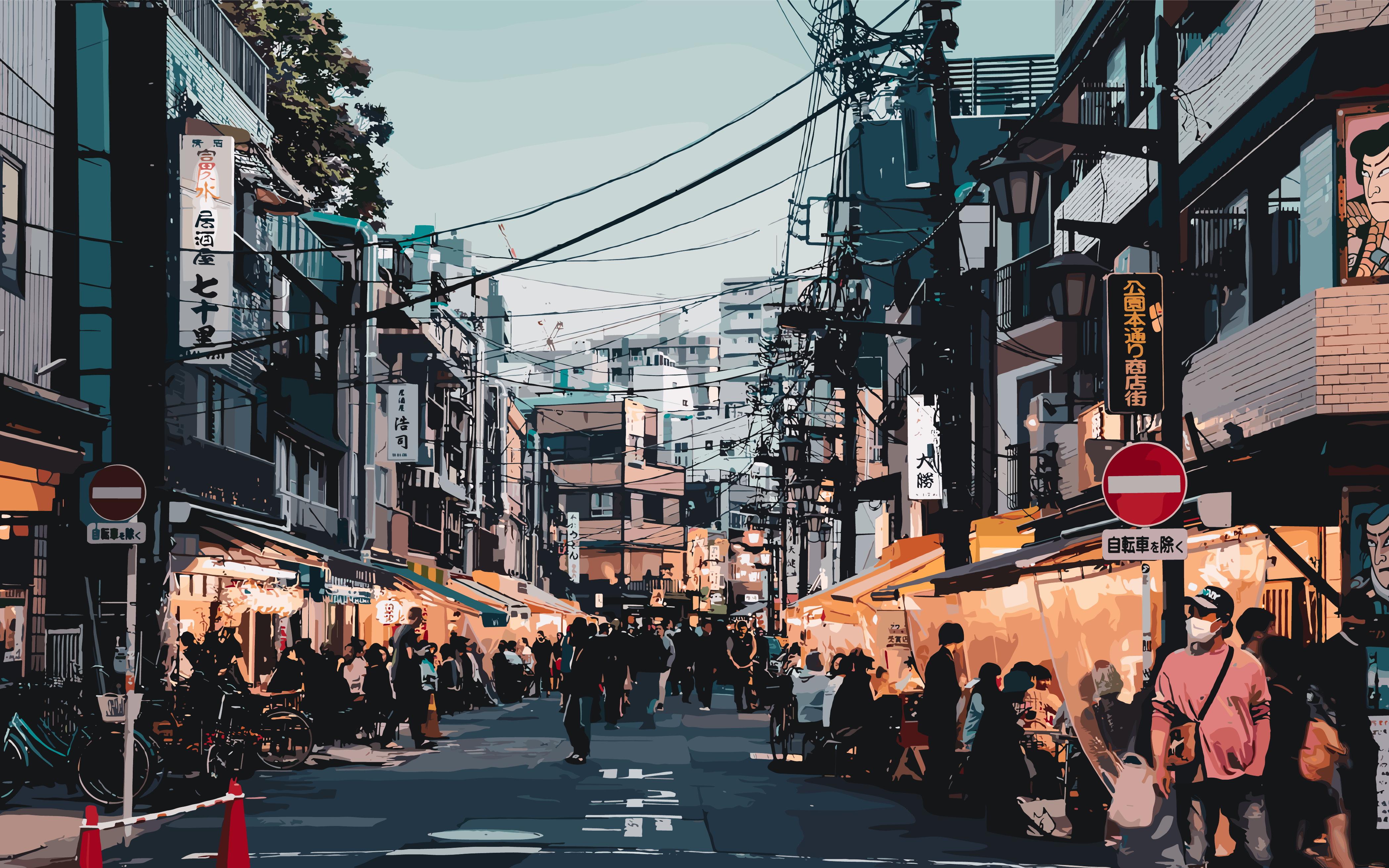}
    \end{minipage}%
    \begin{minipage}{0.3\linewidth}
        \raggedright
        \textbf{Candidate labels}
        \begin{itemize}
            \item \textcolor{red}{house}
            \item tree
            \item \textcolor{red}{street lamp}
            \item \textcolor{red}{people}
            \item mountain
            \item \textcolor{red}{bike}
            \item flower
        \end{itemize}
    \end{minipage}
    \caption{ An example of partial multi-label learning. The
image is partially labeled by noisy annotators. Among the candidate labels, house, street lamp, people and bike are ground-truth labels while tree, mountain and flower
are noisy labels.}
\end{figure}

Formally speaking, let \(\mathcal{D}=\left\{\left(x_i, y_i\right) \mid 1 \leq i \leq n\right\}\) be the partial multi-label training set. \(X=\left\{x_i \mid 1 \leq i \leq n\right\} \in \mathbb{R}^{n \times d}\) which represents \(n\) samples with \(d\)-dimensional feature space and \( Y \in R^{n \times l}\in\{0,1\}^{n \times l}\) be the label space with \(l\) labels. In the label matrix, \(Y_{ik}=1\) means
the \(k\)-th label is one of the candidate labels of \(X_i\). 

The core of addressing the problem of PML is how to deal with ambiguation in 
 label set,  which essentially involves identifying the ground-truth label of an instance from its  candidate label sets. A common idea is to exploit existing information to disambiguate noises in labels. Several methods utilize relationships between labels \cite{xie2018partial,xie2021partial1}, while others focus on analyzing the relationships between labels and features \cite{zhang2020partial,xu2020partial,yu2018feature}. Additionally, some methods combine these two ideas \cite{xie2020semi,sun2021global,li2021partial}.

However, existing research predominantly leverages the information contained within the labels or the relationship between labels and features. While features also contain valuable information that can be exploited. In partial label learning scenarios, the information within features is considered to be relatively accurate in contrast to the potentially erroneous label information. By using the accurate information within the features,  the ground-truth label of a sample can be identified from its candidate set. Moreover, the structural consistency between labels and features has not been thoroughly considered, i.e., similar features correspond to similar labels. By integrating these two characters, we can utilize the accurate information within the features to remove noises from the label set. 

Another problem is the sparsity of multi-label datasets. In multi-label datasets, the number of positive labels is much smaller than the number of negative labels, which naturally interferes with the learning process of the model due to data imbalance. Due to the significant difference in the number of learning examples for positive labels compared to negative labels, the models' ability to distinguish positive labels is weak. However, in practical application scenarios such as fault detection \cite{abid2021review} and disease diagnosis \cite{kumar2023artificial}, the importance of distinguishing positive labels is much higher than that of distinguishing negative labels. Therefore, how to improve the model's ability to distinguish true positive labels  when constructing a model is a key issue.

Feature selection can find the most informative and representative features, remove redundant and irrelevant features, and reduce the curse of dimensionality. During this process, we find that there are some problems in the common used feature selection terms in embedded feature selection: 
certain feature selection methods extend the linear projection of features to labels, resulting in limitations when handling high-order relations. Others rely on latent space coefficients for feature selection, but these methods often lack robustness when dealing with redundancy, erroneous information, and structural inconsistencies.

\begin{figure}[t]
\centering
\includegraphics[width=0.9\columnwidth]{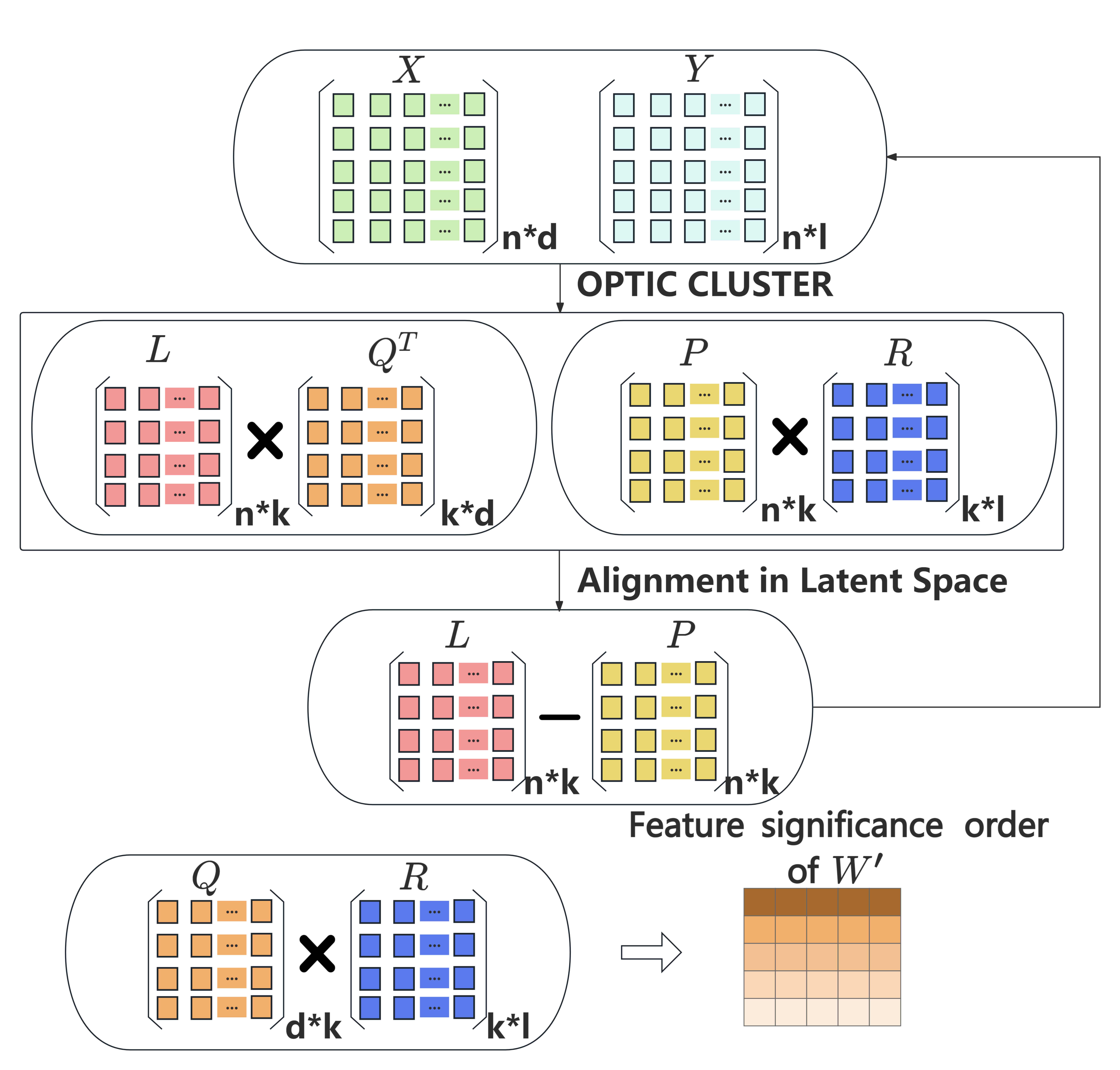} 
\caption{The Process of PML-FSLA. First, the feature matrix and the label matrix are projected into the \(k\)-dimension space determined by OPTICS. Then noisy labels are removed through latent space alignment. Finally two weight matrices are employed for feature selection. }
\label{figure 2}
\end{figure}

To solve these problems, we propose Partial Multi-label Feature Selection based on Latent Space Alignment (PML-FSLA). The whole process is shown in Figure  \ref{figure 2}. First  OPTICS clustering is utilized to identify dimensions of latent space. Then   matrix decomposition is applied to project the label matrix and feature matrix  into a latent space, and the information of features  is employed to reduce the noises of labels through alignment. Finally, feature selection is performed using the projection coefficient matrix of label matrix and feature matrix. Our main contributions are:
\begin{itemize}
\item A new feature selection term is utilized, involving the product of the projection coefficient matrix of the feature matrix and the label matrix. This method, compared to traditional coefficient matrices or single projection coefficient matrices, is more flexible and better suited for handling high-order relationships. The structural consistency of the feature matrix and label matrix is also utilized more directly, enhancing the identification capabilities for true positive labels.
\item The correct information in the feature matrix is leveraged through latent space alignment, while noises in the label matrix are reduced by utilizing the structural consistency between the label matrix and the feature matrix.
\item The OPTICS \cite{ankerst1999optics} clustering method is employed to determine the dimension of the latent space. This allows the model to determine the number of clusters based on the characteristics of the dataset, making the construction of the latent space more reasonable.
\end{itemize}

\section{Related Work}
\subsection{OPTICS Clustering}
OPTICS clustering refers to “Ordering Points To Identify the Clustering Structure”, is a classic clustering algorithm first introduced in 1999 \cite{ankerst1999optics}. It is an extension of DBSCAN (Density-Based Spatial Clustering of Applications with Noise) algorithm \cite{ester1996density}, the main idea is to find the cluster of the dataset by identifying the density-connected points. The algorithm builds a density-based representation of the data by creating an ordered list of points called the reachability plot. Each point in the list is assigned with a reachability distance to measure how far it is to reach that point from other points in the dataset. Points with similar reachability distances are likely to be in the same cluster.

Compared to tradition cluster method like KNN, the number of clusters does not need to be set in advance. It provides more flexibility in selecting the number of clusters, can better identify clusters of arbitrary shapes and effectively handle noises and outliers. Compared to DBSCAN, OPTICS can handle data clusters with varying density and generate richer output and it is less sensitive to noises. 
\subsection{Embedding Method Feature Selection}
In embedded feature selection, a common idea is to take  term $\|XW-Y\|_{F}^2 $  as the core term, then add other regularization terms with different meanings to form the objective function, and obtain the final feature selection term \(W\) by optimizing the objective function and select the feature according to $\|W_{i.}\|_2$. Some methods remove the noises or cluster \(X\)  or \(Y\)  by adding the matrix decomposition term \cite{he2019discriminatively,li2020recovering,yu2020partial}. Hu et. al. 
 propose a method eliminates the noises by
decomposing the label matrix into the low-dimensional space \cite{hao2023partial}. The feature matrix is also decomposed to steer the direction for label disambiguation. Finally, the coherent subspace is constructed through the shared coefficient weight matrix. Some methods add manifold terms for structural consistency of the matrix  decomposition \cite{tibshirani1996regression,qi2018unsupervised,wei2016unsupervised}. Shang et. al. propose a graph regularized feature selection framework based on subspace learning, which is extended by introducing data graphs. Data graph and feature graph are introduced into subspace learning to maintain the geometric structure of data manifold and feature manifold \cite{shang2020sparse}. Some methods add regularization term to ensure the sparsity of the matrix \cite{nie2010efficient,akbari2019elastic}. Benefit from the joint advantages of the dual-manifold learning and the hesitant fuzzy correlation, it adds a feature manifold regularization term based on HFCM between features to the objective function to maintain the similarity between features and feature weights \cite{mokhtia2021dual}. In addition, the regularization term of the sample manifold is also considered to maintain the local correlation between each class of samples.

The main drawback of  linear feature selection term  is to depict the mapping from \(X\) to \(Y\) through low-order linear relationship. However, in high-dimensional and large-scale multi-label data, the high-order relationship between features and labels is more complex, and it is difficult to depict it simply with low-order relationship.

Another common method is to exploit the latent space for feature selection. Some methods project the original matrix into the latent space to reduce noises or redundancy. Braytee et. al.  solve the high-dimensional problem of data by matrix decomposition of label space and feature space respectively \cite{braytee2017multi}. By projecting the original space to the low-dimensional space, it achieves the purpose of distinguishing irrelevant labels, related labels and wrong labels. Others believe that the label matrix and the feature matrix have structural consistency, and are the mapping to different dimensions of the same high-dimensional space \cite{hu2020multi}. These methods fit the projection of the initial matrix in the objective function, and select features through the coefficient matrix corresponding to the projection matrix. By projecting the label matrix and the feature matrix into the same dimension, Gao et. al.  explore the impact of potential feature structure on label relationship, and design a latent shared term to share and preserve both latent feature structure and latent label structure \cite{gao2021multilabel}.

The main disadvantage of single latent feature selection term is that the projection coefficient matrix used means the weight coefficient of the latent space projected to the feature matrix. The premise of feature selection in this way is that the latent space projection of the feature matrix and the label matrix eventually converge, so the corresponding feature weight and label weight also converge. While the problem is that even if the feature matrix and label matrix have structural consistency, the accuracy cannot be guaranteed due to: (1) the existence of redundant information and error information; (2) the structure cannot be completely consistent.

\section{The Proposed Method }

To find the latent space dimensions that are consistent with the dataset, the OPTICS clustering method is first used on the features. The radius is set as \(r\), after which the latent dimension \(k\) is obtained. Then, the feature matrix \(X \in \mathbb{R}^{n \times d}\) is decomposed into two low-dimensional matrices \(L \in \mathbb{R}^{n \times k}\) and \(Q^T \in \mathbb{R}^{k \times d}\). To minimize the reconstruction error, the following form is obtained:

\begin{equation}
    \min _{Q,L}\|X-LQ^T\|_{F}^2.\label{f1}
\end{equation}
Where \(L\) represents the latent cluster matrix of feature matrix, with \(Q^T\) denoting a coefficient matrix. Formula \ref{f1}
indicates that \(d\)-dimensional features reduce to \(k\)-dimensional
features, which is obtained by OPTICS. It can be explained that the original \(d\)-dimensions features are clustered into \(k\) different clusters, relevant features are in the same cluster, while features of different clusters are independent. The original feature matrix \(X\) can be seen as projected from the latent matrix \(L\), and \(Q^T\) is the weight coefficient matrix. The row of matrix \(Q\) represents the
coefficient of each feature in these \(k\) latent feature variables.

\begin{figure}[htbp]
\centering
\includegraphics[width=0.9\columnwidth]{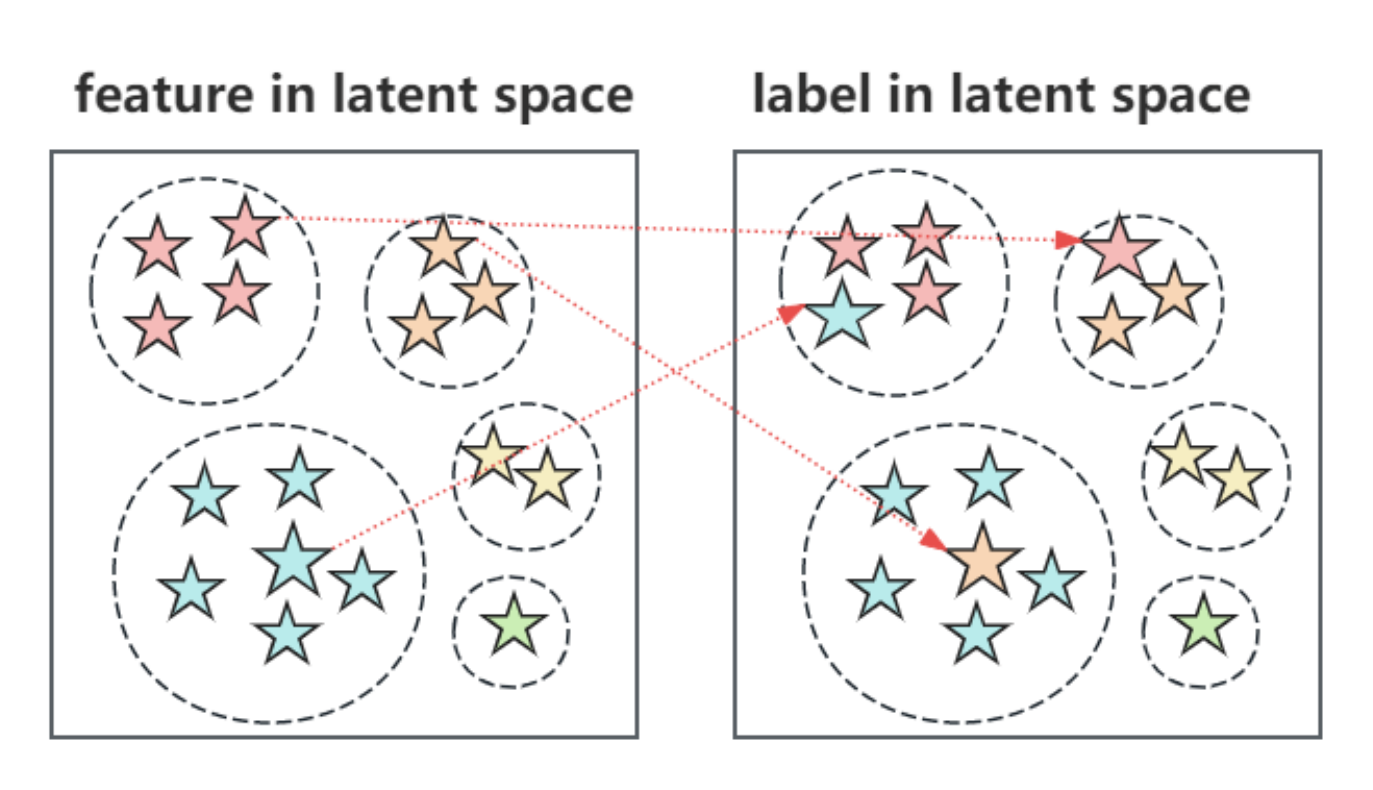} 
\caption{Through alignment of labels and features in latent space, noisy labels can be found and eliminated due to the structural inconsistencies}
\label{figure 3}
\end{figure}

Similarly, label matrix \(Y\) can also be decomposed into latent cluster matrix \(P\)  and coefficient matrix \(R\). As the correct label matrix \(T\) should have a strong corresponding relationship with the characteristic matrix, the projection of the feature matrix and the label matrix should be consistent in the latent space. In addition, in the partial multi-label scenario, the label matrix is considered correct and noises appears in the label matrix. Combining these two points, we can utilize the latent space alignment of the feature matrix and the label matrix to reduce the noises in the label by using the correct information in the feature matrix. As shown in Figure \ref{figure 3}, by comparing the projection of the label matrix and the feature matrix in the latent space, the label projection matrix can be updated according to the feature projection matrix:
\begin{equation}
    \min _{Q,L,P,R,T}\|X-LQ^T\|_{F}^2+\alpha\|T-PR\|_{F}^2+\beta\|L-P\|_{F}^2 .
\end{equation}
Where \(\alpha\) and \(\beta\) are parameters employed to balance the contribution of label decomposition and latent space alignment. \(T\) is the correct label matrix, and its initial value is \(Y\). The third term is the alignment term to match the  projections of the feature matrix and the label matrix so that the correct information of the feature matrix can be exploited. Unlike the traditional feature selection using \( W\in R^{d \times l}\) where \(d\), \(l\) are the dimension of feature and label as the feature selection term, this method adopts the product of \(Q\) and \(R\) as the feature selection term, so the final function is reformed as:
\begin{equation}
\begin{aligned}
    &\min_{Q,L,P,R,T} \|X - LQ^T\|_F^2 + \alpha\|T - PR\|_F^2 +
    \beta\|L - P\|_F^2 \\&+\gamma\|QR\|_{2,1},\\
    & s.t. Q,L,P,R,T \geq 0.\label{f3}
\end{aligned}
\end{equation}
Where \(\theta\) is a parameter that ensures the sparsity of the objective function and carry out feature selection. In the \(QR\) method, \(Q\) represents the coefficients of the clustering \(L\) projected onto \(X\), while \(R\) represents the coefficients of the clustering \(T\) projected onto \(Y\). With the iterations progress, the latent projections of the feature matrix and the label matrix tend to converge due to the alignment term's influence. \(QR\) thus indicates the similarity of the same class \(K\) in terms of their projections onto \(X\) and \(Y\), \(QR_{ij}\) reflects the similarity between a feature \( X_i \) and a label \(Y_j \) in latent space. The higher the \(QR_{ij}\), the more similar \( X_i \)  and \(Y_j \) are in latent space, the more important \( X_i \) is to \( Y_j \).
Compared to commonly used feature selection term \( W \), the starting point of this method is to deploy the structural consistency between labels and features to project to the common dimension \( k\) for comparison and then select features.  And it is more flexible for processing high-order relationships as traditional $\|XW-Y\|_{F}^2 $ is relatively linear.

Compared to another commonly used feature selection term \( Q \), the meaning of each column is the coefficient of cluster matrix projected onto \( X \). The premise of feature selection in this method is that the learned \( L \) and \( P \) eventually converge. Therefore, the feature weight of \( L \) for \( X \) and the label weight of \( P \) for \( Y \) also converge. However, the problem is that even though \( X \) and \( Y \) have structural consistency, the accuracy cannot be guaranteed due to the existence of redundant information and noises and the structure cannot be completely consistent. By incorporating \( R \), which is related to \( Y\), this method can effectively reduce the impact of redundancy and noises, and mitigate potential inconsistencies within the structure. 

Moreover, incorporating \( R \) in the \( QR \) method inherently enhances the ability to identify true positive labels during feature selection. Since \( X,Y,V,Q,L,P,T \) are all non-negative, in  if \(Y_{ij}\) is equal to zero, then the coefficient of the latent matrix must be zero. In other words, if \( P_{ij} \) is not zero, the corresponding y is not equal to zero. Therefore, in \( QR \), increasing of the weight is only related to positive labels. The features selected according to the \( QR \) weight are more consistent with positive labels in latent space, which strengthens the  identification of  positive labels.
\begin{algorithm}[h]
    \caption{Pseudo code of PML-FSLA}
    \label{alg:algorithm}
    \textbf{Input}:\\Feature matrix \(X\) and label matrix \(Y\);\\Regularization parameters \(\alpha\), \(\beta\), and \(\gamma\);
    \\Cluster radius \(r\).\\
    \textbf{Output}:Return the ranked features.
    
    \begin{algorithmic}[1] 
        \STATE Cluster \(X\) with the given radius \(r\) using the OPTICS  and determine the cluster number \(k\);
        \WHILE{not coverage}
        \STATE Update \(L\) by Formula \ref{f17} with other variables fixed;
        \STATE Update \(Q\)  by Formula \ref{f16} with other variables fixed;
        \STATE Update \(P\)  by Formula \ref{f18} with other variables fixed;
        \STATE Update \(R\)  by Formula \ref{f19} with other variables fixed;
        \STATE Update \(T\)  by Formula \ref{f20} with other variables fixed;
        \ENDWHILE
        \STATE \textbf{return} Return features according to \(\left\|QR_i\right\|_2\).
    \end{algorithmic}
\end{algorithm}

\section{Optimization }
As the Formula \ref{f3} is joint nonconvex,  the global optimal solution cannot be obtained. Moreover, the Formula \ref{f3}  is nonsmooth due to the existence of \(l_{2,1}\)-norm feature selection term. Therefore, three iterative rules are introduced to obtain the local optimal solution. First  \( QR \) is relaxed as  \(\operatorname{Tr}(QR)^T D(QR)\) where \(D \in R^{d \times d}\) is a diagonal matrix and the \( i \)-th diagonal element \( D_{i i}=\left(1 /\left(2\left\|QR_{i \cdot}\right\|_2+\epsilon\right)\right) \).  \( \epsilon \) is a extremely small positive constant. So the Formula  \ref{f3}  can be rewritten as:
    
\begin{equation}
    \begin{aligned}
\Theta(Q, L, P, R,T)= & \operatorname{Tr}\left(\left(X^T-Q L^T\right)\left(X-L Q^T\right)\right) \\
& +\alpha \operatorname{Tr}\left(\left(T^T-R^T P^T\right)(T-PR)\right) \\
& +\beta \operatorname{Tr}\left(\left(L^T-P^T\right)(L-P)\right)\\&+2 \gamma \operatorname{Tr}\left(R^TQ^T D QR\right).\label{f4}
\end{aligned}
\end{equation}

After that nonnegative constraints are integrated into Formula \ref{f4} and the lagrange function of it can be obtained:
\begin{equation}
\begin{aligned}
& \mathcal{L}(Q, L, P, R,T) \\
 =&\operatorname{Tr}\left(\left(X^T-Q L^T\right)\left(X-L Q^T\right)\right) \\
& +\alpha \operatorname{Tr}\left(\left(T^T-R^T P^T\right)(T-PR)\right) \\
& +\beta \operatorname{Tr}\left(\left(L^T-P^T\right)(L-P)\right)\\&+2 \gamma \operatorname{Tr}\left(R^TQ^T D QR\right)-\operatorname{Tr}\left(\Omega Q^T\right)-\operatorname{Tr}\left(\Psi L^T\right)\\&
-\operatorname{Tr}\left(\Phi P^T\right)-\operatorname{Tr}\left(\Upsilon R^T\right)-
\operatorname{Tr}\left(\tau T^T\right).
\end{aligned}
\end{equation}
Where \(\Omega\in R^{d \times k}\), \(\Psi\in R^{n \times k}\), \(\Phi\in R^{n \times k}\), \(\Upsilon\in R^{k \times l}\) and \(\tau \in R^{n \times l}\) are
Lagrange multipliers. The partial derivatives of the function w.r.t variables \( Q\), \( L \), \( P \), \( R \) and \( T \) are:
    \begin{equation}
        \frac{\partial \mathcal{L}}{\partial Q}=-2 X^T L+2 QL^TL+2 \gamma QRR^T -\Omega.
    \end{equation}
    \begin{equation}
        \frac{\partial \mathcal{L}}{\partial L}=-2 X Q+2 L Q^T Q +2 \beta L- 2 \beta P- \Psi.
    \end{equation}
    \begin{equation}
        \frac{\partial \mathcal{L}}{\partial P}=-2\alpha T R^T+2\alpha PR^T R +2 \beta P- 2 \beta L- \Phi.
    \end{equation}
    \begin{equation}
        \frac{\partial \mathcal{L}}{\partial R}=-2\alpha P^T T+2\alpha P^TPR+2 \gamma QQ^TR -\Upsilon. 
    \end{equation}
    \begin{equation}
        \frac{\partial \mathcal{L}}{\partial T}=2\alpha T- 2\alpha PR -\tau.
    \end{equation}
Based on Karush–Kuhn–Tucker condition  we obtain that:
\begin{equation}
        (-2 X^T L+2 QL^TL+2 \gamma QRR^T )_{ij}\Omega_{ij}=0.
    \end{equation}
\begin{equation}
        (-2 X Q+2 L Q^T Q +2 \beta L- 2 \beta P)_{ij}\Psi_{ij}=0.
    \end{equation}
\begin{equation}
        (-2\alpha T R^T+2\alpha PR^T R +2 \beta P- 2 \beta L)_{ij}\Phi_{ij}=0.
    \end{equation}
\begin{equation}
        (-2\alpha P^T T+2 \alpha P^TPR+2 \gamma QQ^TR  )_{ij}\Upsilon_{ij}=0.
    \end{equation}
    \begin{equation}
        (2\alpha T- 2\alpha PR)_{ij}\tau_{ij}=0.
    \end{equation}
Then \( Q\), \( L \), \( P \), \( R \) and \( T \) can be presented as:
\begin{equation}
    L_{i j}^{t+1} \leftarrow L_{i j}^t \frac{\left(X Q+ \beta P\right)_{i j}}{\left(L Q^T Q +\beta L\right)_{i j}}.\label{f17}
    \end{equation}
\begin{equation}
    Q_{i j}^{t+1} \leftarrow Q_{i j}^t \frac{\left(  X^TL\right)_{i j}}{\left( QL^TL+ \gamma QRR^T\right)_{i j}}.\label{f16}
    \end{equation}
\begin{equation}
    P_{i j}^{t+1} \leftarrow P_{i j}^t \frac{\left( \alpha T R^T+2 \beta L\right)_{i j}}{\left(\alpha PR^T R +\beta P\right)_{i j}}.\label{f18}
    \end{equation}
\begin{equation}
    R_{i j}^{t+1} \leftarrow R_{i j}^t \frac{\left(\alpha P^T T \right)_{i j}}{\left( \alpha P^TPR+\gamma QQ^TR\right)_{i j}}.\label{f19}
    \end{equation}
\begin{equation}
    T_{i j}^{t+1} \leftarrow T_{i j}^t \frac{\left(PR)\right)_{i j}}{\left( T\right)_{i j}}.\label{f20}
    \end{equation}
Where \(t\) indicates the iterative number. To ensure that the denominator does not reach zero during model iteration, a very small constant is added to the denominator of each term. The whole process is presented in Algorithm \ref{alg:algorithm}.

\section{Experiment}
\subsection{Experimental Setup}

To prove the effectiveness of the proposed method, we compare our method with the nine state-of-the-art methods of partial multi-label learning and multi-label learning method. Six partial multi-label learning methods (PML-LC \cite{xie2018partial}, PML-FP \cite{xie2018partial}, PAR-VLS \cite{zhang2020partial}, PAR-MAP \cite{zhang2020partial}, FPML \cite{yu2018feature} and PML-FSSO \cite{hao2023partial}) and three multi-label feature selection methods (MLKNN \cite{2007ML}, MIFS \cite{2016Multi} and DRMFS \cite{2020Robust}) are involved.  One classical  and two low-rank  multi-label feature selection methods are adopted. Due to the lack of feature selection method in partial multi-label learning,  the weight matrix is extracted from  model to reflect the importance of features. We adopt
ten-fold cross-validation to train these models and the selected features are compared on SVM classifier. 
\subsection{Datasets and Evaluation Metrics}
We conduct experiments on eight datasets from multiple fields: \textbf{\textit{HumanPseAAC}} for Biology, \textbf{\textit{CAL500}} for music classification, \textbf{\textit{Chess}} and \textbf{\textit{Corel5K}}  for image
annotation, \textbf{\textit{LLOG\_F}} and for  text categorization, \textbf{\textit{Water}} for chemistry, \textbf{\textit{Yeast}} for gene function prediction and \textbf{\textit{CHD49}}
for medicine. Detailed information of datasets is shown in Table \ref{t1}.
\begin{table}[h]
  \centering
  {\scriptsize 
    \begin{tabular}{@{}lllll@{}}
      \toprule
      Name & Domain & \#Instances & \#Features & \#Labels \\ \midrule
      CAL         & music     & 555  & 49    & 6    \\
      CHD\_49     & medicine  & 555  & 49    & 6    \\
      Chess       & imagine   & 585  & 258   & 15   \\
      Corel5K     & image     & 5000 & 499   & 374  \\ 
      HumanPseAAC & biology   & 3106 & 40    & 14   \\ 
      LLOG\_F     & text      & 1460 & 1004  & 75   \\
      Water       & chemistry & 1060 & 16    & 14   \\
      Yeast       & biology   & 2417 & 103   & 14   \\ \bottomrule
    \end{tabular}
  }
  \caption{Characteristics of experimental datasets.}
  \label{t1}
\end{table}

\subsection{Results}
\begingroup
\setlength{\tabcolsep}{5pt}
\begin{table*}[htbp]

{\scriptsize
\begin{tabular}{@{}lllllllllll@{}}
\toprule
Datasets    & PML-FSLA               & PML-LC        & PML-FP        & PAR-VLS       & PAR-MAP       & FPML        & PML-FSSO      & MLKNN         & MIFS          & DRMFS         \\ \midrule
CAL         & \textbf{0.538±0.019} & 0.018±0.031   & 0.069±0.049   & 0.020±0.001    & 0.079±0.078   & 0.122±0.070  & 0.000±0.000        & 0.043±0.043   & 0.000±0.000           & 0.000±0.000           \\
CHD\_49     & \textbf{0.637±0.059} & 0.310±0.149    & 0.244±0.118   & 0.202±0.136   & 0.262±0.120    & 0.012±0.014 & 0.484±0.160  & 0.310±0.188    & 0.157±0.124   & 0.090±0.127    \\
Chess       & \textbf{0.579±0.061} & 0.404±0.082 & 0.432±0.128 & 0.497±0.127 & 0.536±0.145 & 0.050±0.000    & 0.035±0.003 & 0.552±0.147 & 0.299±0.081 & 0.154±0.027 \\
Corel5K     & \textbf{0.263±0.094} & 0.075±0.020    & 0.067±0.047   & 0.054±0.058   & 0.043±0.016   & 0.000±0.000      & 0.004±0.004 & 0.000±0.000           & 0.000±0.000          & 0.035±0.039   \\
LLOG F      & \textbf{0.339±0.006} & 0.178±0.056   & 0.162±0.034   & 0.033±0.000       & 0.230±0.123    & 0.315±0.085 & 0.081±0.035 & 0.305±0.127   & 0.187±0.065   & 0.024±0.006   \\
HumanPseAAC & \textbf{0.185±0.002} & 0.011±0.018 & 0.074±0.022 & 0.145±0.091 & 0.188±0.114 & 0.000±0.000        & 0.008±0.001 & 0.056±0.025 & 0.053±0.042 & 0.000±0.000          \\
Water       & \textbf{0.556±0.000}     & 0.331±0.149   & 0.346±0.176   & 0.094±0.051   & 0.151±0.106   & 0.106±0.073 & 0.186±0.139 & 0.149±0.137   & 0.455±0.018   & 0.056±0.088   \\
Yeast       & \textbf{0.496±0.022} & 0.000±0.000          & 0.000±0.000          & 0.000±0.000           & 0.000±0.000            & 0.000±0.000        & 0.026±0.025&0.020±0.026    & 0.228±0.159   & 0.021±0.005   \\ \bottomrule
\end{tabular}
}

\caption{ Experimental results (mean ± std) in terms of Mirco-f1 where the best performance is shown in boldface.}
 \label{t2}
\end{table*}

\begin{table*}[htbp]
{\scriptsize
\begin{tabular}{@{}lllllllllll@{}}
\toprule
Datasets    & PML-FSLA               & PML-LC        & PML-FP        & PAR-VLS       & PAR-MAP       & FPML          & PML-FSSO      & MLKNN         & MIFS          & DRMFS         \\ \midrule
CAL         & \textbf{0.538±0.019} & 0.017±0.029   & 0.074±0.055   & 0.017±0.000       & 0.075±0.075   & 0.104±0.062   & 0.033±0.026 & 0.041±0.042   & 0.000±0.000           & 0.000±0.000            \\
CHD\_49     & \textbf{0.615±0.066} & 0.272±0.131   & 0.226±0.112   & 0.164±0.116   & 0.237±0.118   & 0.011±0.013   & 0.187±0.133 & 0.277±0.179   & 0.134±0.104   & 0.076±0.107   \\
Chess       & \textbf{0.588±0.042} & 0.050±0.000      & 0.343±0.088 & 0.405±0.141 & 0.476±0.139 & 0.131±0.016 & 0.020±0.000              & 0.517±0.156 & 0.539±0.147 & 0.511±0.154 \\
Corel5K     & \textbf{0.231±0.094} & 0.064±0.016   & 0.045±0.034   & 0.041±0.042   & 0.030±0.011    & 0.000±0.000          & 0.003±0.002 & 0.000±0.000             & 0.000±0.000           & 0.022±0.024   \\
LLOG F      & \textbf{0.321±0.005} & 0.027±0.113   & 0.047±0.050    & 0.033±0.000       & 0.227±0.114   & 0.221±0.058   & 0.104±0.037 & 0.273±0.100     & 0.148±0.060    & 0.019±0.005   \\
HumanPseAAC & \textbf{0.185±0.002} & 0.009±0.015 & 0.067±0.019 & 0.125±0.085 & 0.189±0.115 & 0.000±0.000           & 0.004±0.000     & 0.042±0.019 & 0.035±0.028 & 0.000±0.000         \\
Water       & \textbf{0.551±0.000}     & 0.324±0.146   & 0.337±0.172   & 0.104±0.056   & 0.167±0.112   & 0.111±0.074   & 0.134±0.091 & 0.149±0.142   & 0.444±0.015   & 0.065±0.092   \\
Yeast       & \textbf{0.488±0.022} & 0.000±0.000             & 0.000±0.000             & 0.000±0.000             & 0.000±0.000  & 0.000±0.000            & 0.029±0.026 & 0.019±0.025   & 0.203±0.150    & 0.032±0.008   \\ \bottomrule
\end{tabular}
}
\caption{ Experimental results (mean ± std) in terms of Marco-f1 where the best performance is shown in boldface.}
 \label{t3}
\end{table*}
\begin{table*}[htbp]
{\scriptsize
\begin{tabular}{@{}lllllllllll@{}}
\toprule
Datasets    & PML-FSLA               & PML-LC        & PML-FP        & PAR-VLS              & PAR-MAP       & FPML          & PML-FSSO               & MLKNN         & MIFS          & DRMFS         \\ \midrule
CAL         & \textbf{0.583±0.004} & 0.388±0.008   & 0.407±0.013   & 0.476±0.045          & 0.545±0.040    & 0.554±0.049   & 0.554±0.053          & 0.541±0.041   & 0.468±0.075   & 0.429±0.038   \\
CHD\_49     & 0.712±0.010           & 0.683±0.007   & 0.663±0.008   & 0.763±0.023          & 0.713±0.007   & 0.646±0.020    & \textbf{0.773±0.035} & 0.739±0.018   & 0.748±0.024   & 0.765±0.031   \\
Chess       & \textbf{0.761±0.037} & 0.543±0.085 & 0.546±0.093 & 0.616±0.117        & 0.663±0.133 & 0.253±0.033 & 0.714±0.133          & 0.666±0.124 & 0.476±0.129 & 0.426±0.122 \\
Corel5K     & \textbf{0.413±0.061} & 0.302±0.034   & 0.315±0.039   & 0.285±0.040           & 0.239±0.020    & 0.266±0.015   & 0.374±0.082          & 0.266±0.025   & 0.282±0.051   & 0.333±0.066   \\
LLOG F      & \textbf{0.591±0.000}     & 0.456±0.004   & 0.478±0.005   & 0.484±0.005          & 0.565±0.028   & 0.600±0.036     & 0.462±0.012          & 0.539±0.027   & 0.542±0.026   & 0.473±0.004   \\
HumanPseAAC & \textbf{0.425±0.000}     & 0.333±0.064 & 0.358±0.048 & 0.447±0.117        & 0.407±0.088 & 0.309±0.046 & 0.347±0.063          & 0.336±0.051 & 0.280±0.070   & 0.337±0.072 \\
Water       & 0.566±0.003          & 0.518±0.009   & 0.530±0.008    & \textbf{0.602±0.013} & 0.591±0.023   & 0.607±0.019   & 0.576±0.044          & 0.588±0.041   & 0.525±0.027   & 0.552±0.046   \\
Yeast       & \textbf{0.690±0.000}      & 0.506±0.031   & 0.521±0.029   & 0.613±0.043          & 0.570±0.006    & 0.633±0.041   & 0.574±0.042          & 0.597±0.040    & 0.678±0.077   & 0.578±0.051   \\ \bottomrule
\end{tabular}
}
\caption{ Experimental results (mean ± std) in terms of Average Precision where the best performance is shown in boldface.}
 \label{t4}
\end{table*}
\begin{table*}[htbp]
{\scriptsize
\begin{tabular}{@{}lllllllllll@{}}
\toprule
Datasets    & PML-FSLA               & PML-LC        & PML-FP        & PAR-VLS       & PAR-MAP       & FPML          & PML-FSSO               & MLKNN                & MIFS          & DRMFS         \\ \midrule

CAL         & \textbf{0.318±0.002} & 0.629±0.120    & 0.613±0.152   & 0.687±0.134   & 0.716±0.067   & 0.684±0.088   & 0.394±0.098          & 0.605±0.093          & 0.579±0.255   & 0.574±0.174   \\
CHD\_49     & 0.327±0.046          & 0.463±0.155   & 0.470±0.133    & 0.339±0.143   & 0.609±0.057   & 0.720±0.082    & \textbf{0.314±0.186} & 0.337±0.132          & 0.347±0.116   & 0.326±0.172   \\
Chess       & \textbf{0.150±0.053}  & 0.258±0.154 & 0.262±0.164 & 0.453±0.136 & 0.398±0.155 & 0.873±0.039 & 0.223±0.180           & 0.395±0.143        & 0.543±0.183 & 0.554±0.199 \\
Corel5K     & \textbf{0.440±0.159}  & 0.652±0.124   & 0.638±0.122   & 0.843±0.082   & 0.896±0.063   & 0.924±0.022   & 0.626±0.192          & 0.899±0.047          & 0.757±0.145   & 0.708±0.160    \\
LLOG F      & \textbf{0.269±0.000}     & 0.763±0.024          & 0.692±0.016          & 0.806±0.051   & 0.669±0.106   & 0.533±0.176   & 0.446±0.108          & 0.530±0.195           & 0.501±0.206   & 0.630±0.197    \\
HumanPseAAC & \textbf{0.249±0.000}     & 0.451±0.175 & 0.489±0.152 & 0.463±0.277 & 0.608±0.167 & 0.848±0.070  & 0.493±0.249          & 0.802±0.082        & 0.733±0.239 & 0.586±0.241 \\
Water       & 0.387±0.001          & 0.458±0.055   & 0.436±0.056   & 0.483±0.028   & 0.440±0.029    & 0.443±0.033   & 0.388±0.063          & \textbf{0.384±0.073} & 0.441±0.033   & 0.418±0.071   \\
Yeast       & \textbf{0.222±0.000}     & 0.432±0.005          & 0.453±0.018          & 0.339±0.137   & 0.623±0.013   & 0.339±0.123   & 0.354±0.136          & 0.339±0.138          & 0.245±0.101   & 0.373±0.158   \\ \bottomrule
\end{tabular}
}
\caption{ Experimental results (mean ± std) in terms of Ranking Loss where the best performance is shown in boldface.}
 \label{t5}
\end{table*}

\begin{table*}[htbp]
{\scriptsize
\begin{tabular}{@{}lllllllllll@{}}
\toprule
Datasets    & PML-FSLA               & PML-LC        & PML-FP        & PAR-VLS       & PAR-MAP              & FPML          & PML-FSSO               & MLKNN         & MIFS          & DRMFS         \\ \midrule
CAL         & \textbf{0.653±0.000}     & 0.807±0.006   & 0.792±0.006   & 0.739±0.034   & 0.714±0.024          & 0.709±0.028   & 0.659±0.039          & 0.703±0.027   & 0.737±0.062   & 0.724±0.042   \\
CHD\_49     & 0.523±0.014          & 0.620±0.038    & 0.622±0.032   & 0.657±0.016   & 0.657±0.016          & 0.666±0.014   & \textbf{0.485±0.072} & 0.532±0.043   & 0.528±0.041   & 0.492±0.069   \\
Chess       & \textbf{0.137±0.017} & 0.200±0.062   & 0.206±0.062 & 0.258±0.060  & 0.233±0.068        & 0.448±0.017 & 0.154±0.082          & 0.230±0.064  & 0.317±0.071 & 0.322±0.085 \\
Corel5K     & \textbf{0.364±0.069} & 0.507±0.030    & 0.483±0.032   & 0.551±0.025   & 0.585±0.006          & 0.574±0.005   & 0.426±0.084          & 0.562±0.016   & 0.510±0.069     & 0.472±0.067   \\
LLOG F      & \textbf{0.536±0.000}     & 0.575±0.077   & 0.583±0.011   & 0.591±0.002   & 0.563±0.011          & 0.550±0.017    & 0.592±0.018          & 0.563±0.015   & 0.568±0.019   & 0.593±0.002   \\
HumanPseAAC & \textbf{0.253±0.000}     & 0.355±0.063 & 0.394±0.030  & 0.286±0.101 & 0.371±0.052        & 0.468±0.011 & 0.339±0.073          & 0.465±0.011 & 0.415±0.082 & 0.362±0.073 \\
Water       & 0.727±0.000              & 0.764±0.010    & 0.747±0.014   & 0.731±0.002   & \textbf{0.726±0.007} & 0.738±0.004   & 0.734±0.038          & 0.733±0.016   & 0.735±0.014   & 0.746±0.032   \\
Yeast       & \textbf{0.520±0.000}      & 0.562±0.080    & 0.576±0.015   & 0.555±0.047   & 0.782±0.000              & 0.532±0.043   & 0.579±0.046          & 0.549±0.049   & 0.522±0.068   & 0.578±0.058   \\ \bottomrule
\end{tabular}
}
\caption{ Experimental results (mean ± std) in terms of Coverage where the best performance is shown in boldface.}
 \label{t6}
\end{table*}
\endgroup
\begin{figure}[htbp]
	\centering
    \begin{subfigure}{\linewidth}
		\centering
		\includegraphics[width=\linewidth]{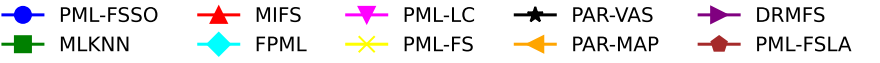}

		\label{ch}
	\end{subfigure}
	\begin{subfigure}{0.49\linewidth}
		\centering
		\includegraphics[width=\linewidth]{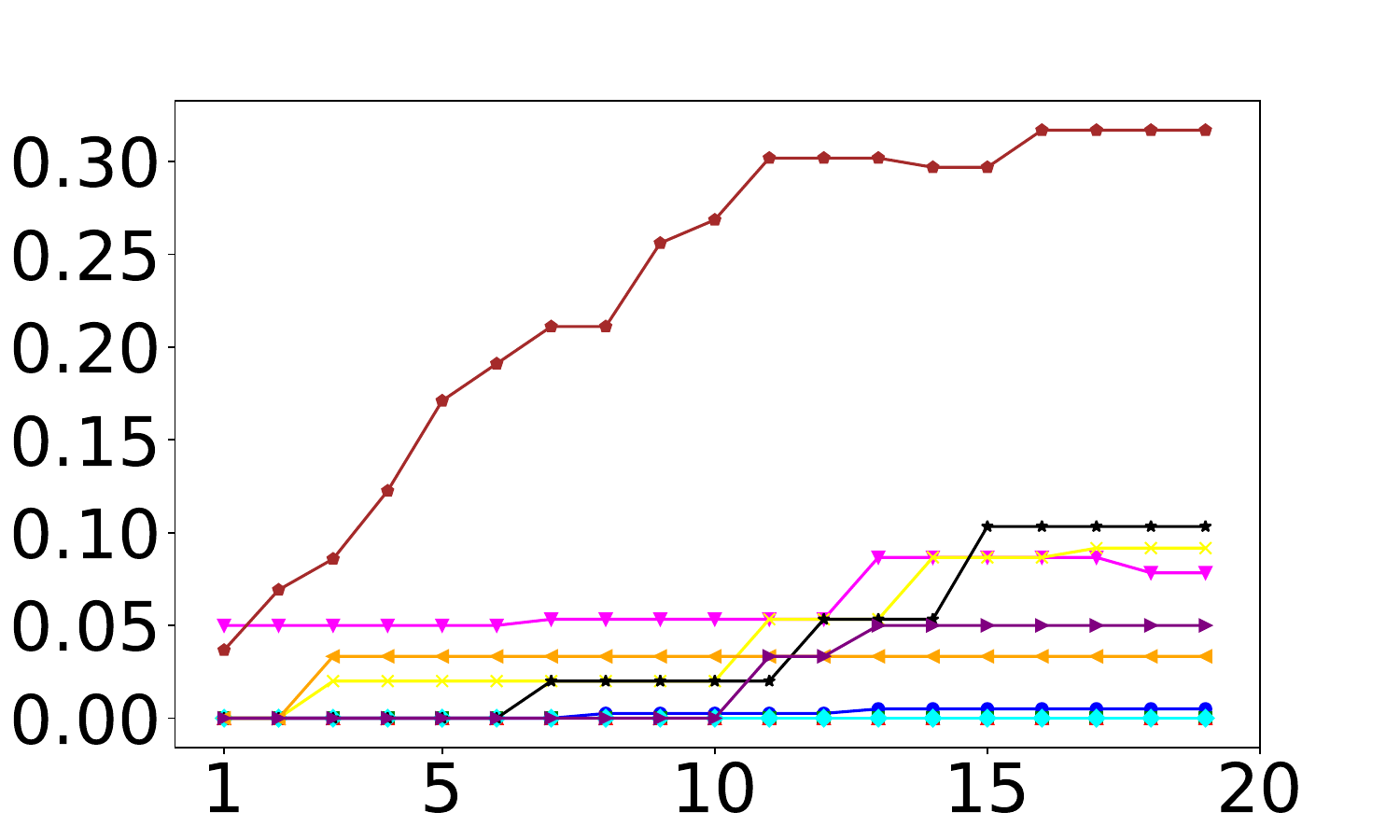}
		\caption{Marco-F1}
		\label{chutian1}
	\end{subfigure}
	\begin{subfigure}{0.49\linewidth}
		\centering
		\includegraphics[width=\linewidth]{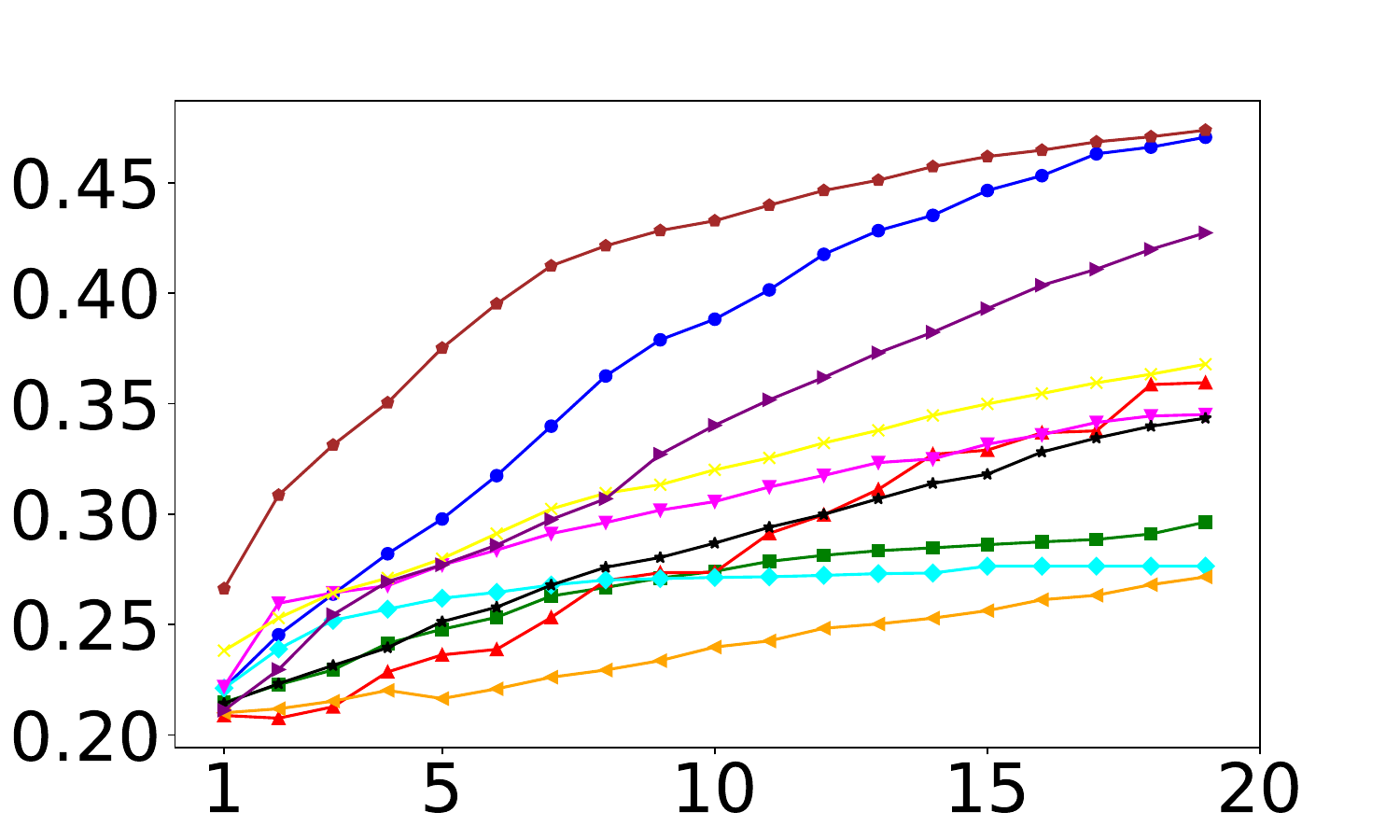}
		\caption{Average
Precision}
		\label{chutian2}
	\end{subfigure}
	
	\begin{subfigure}{0.49\linewidth}
		\centering
		\includegraphics[width=\linewidth]{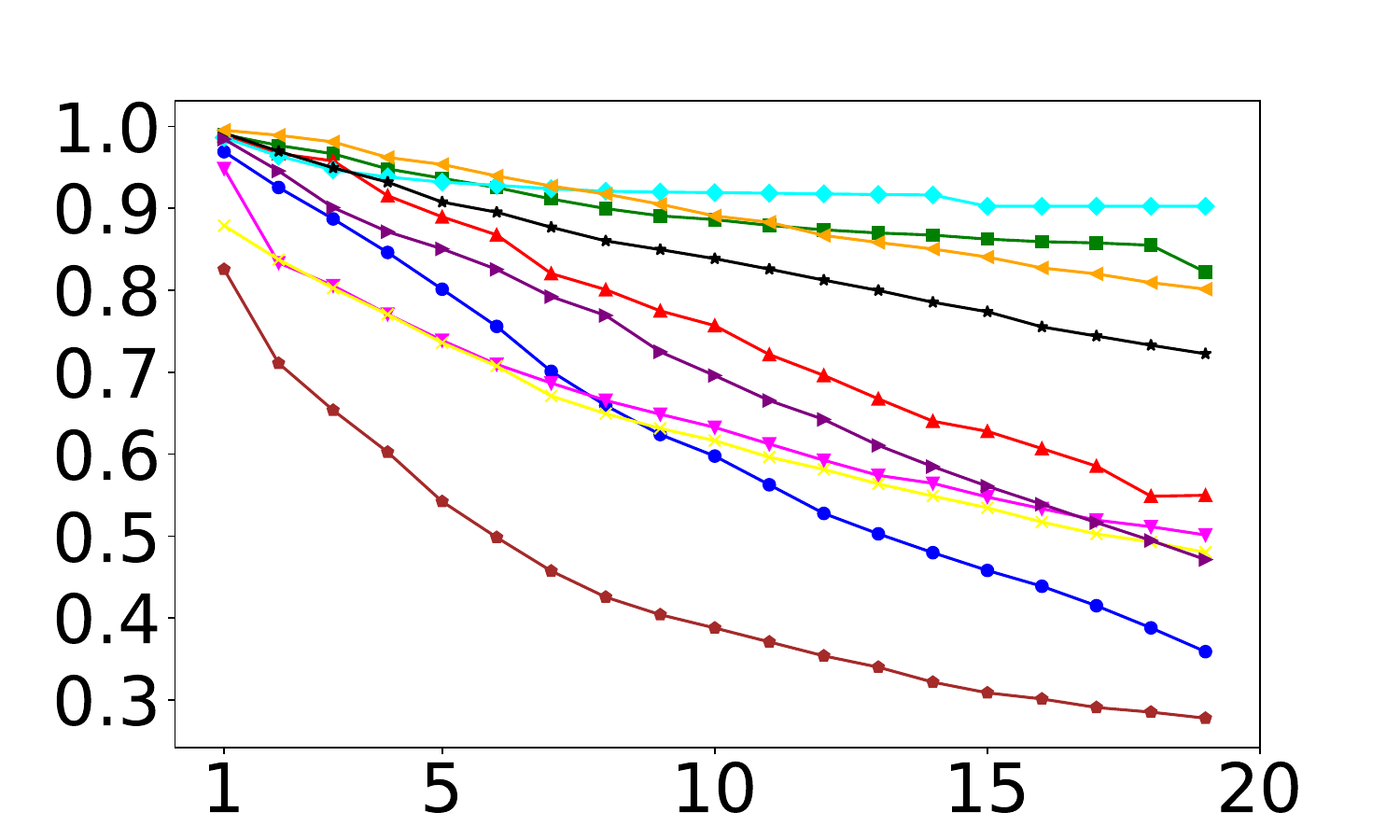}
		\caption{Ranking Loss}
		\label{chutian3}
	\end{subfigure}
	\begin{subfigure}{0.49\linewidth}
		\centering
		\includegraphics[width=\linewidth]{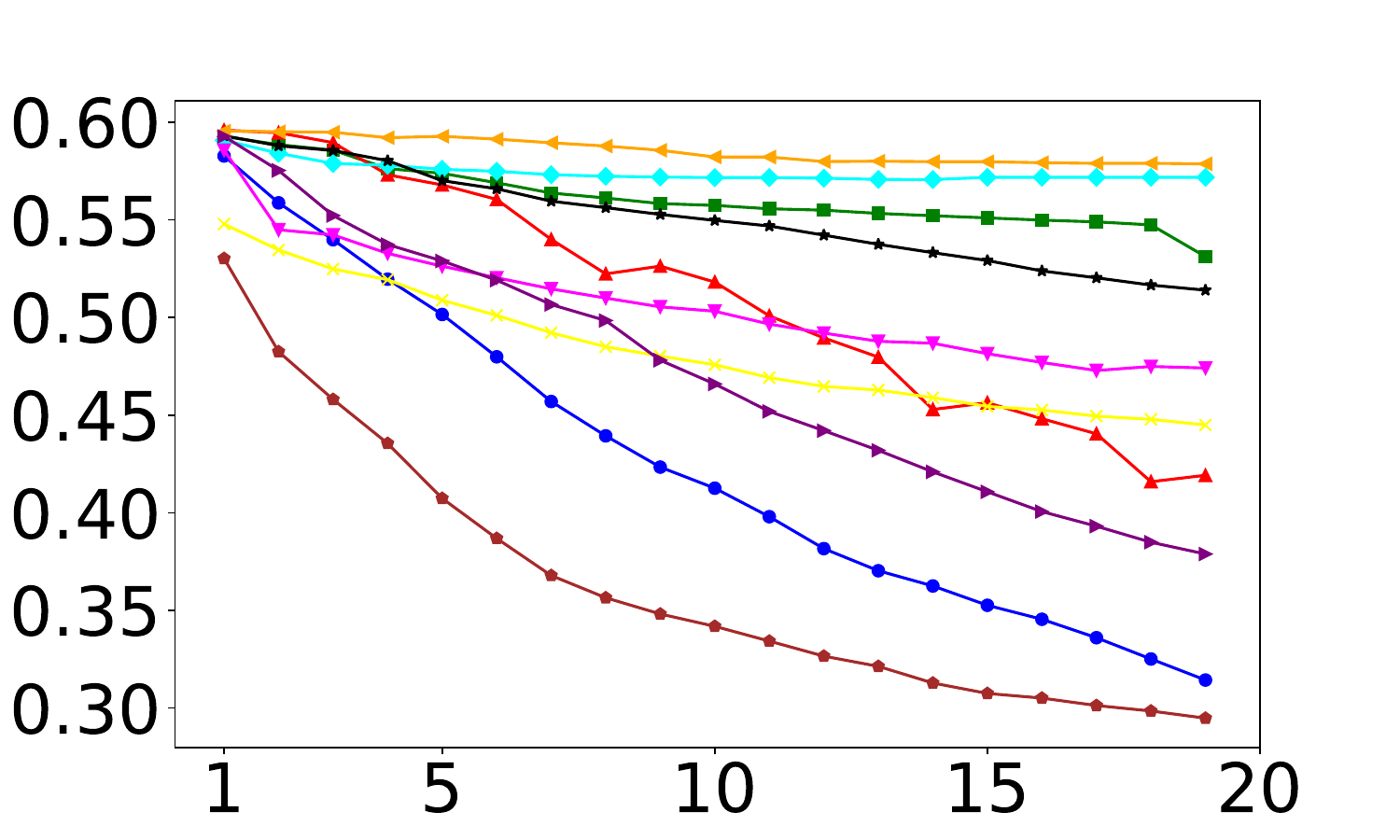}
		\caption{Coverage Error}
		\label{chutian4}
	\end{subfigure}
 \caption{Ten methods on \textbf{\textit{Corel5K}} in terms of Marco-F1, Average
Precision, Ranking Loss and Coverage Error.}
	\label{f3}
\end{figure}
Four widely-used metrics are selected for our performance evaluation: Ranking loss, Coverage, Average Precision, Macro-F1 and Micro-F1. Ranking Loss and Coverage are optimized when their values are minimized, while higher values of Average Precision, Macro-F1 and Micro-F1 indicate better performance.

Tables \ref{t2}-\ref{t6} show the detail of the experiment result, PML-FSLA is adopted  as the abbreviation of our method. For all the datasets except \textbf{\textit{Water}},   one
to twenty percent features  are selected according to the importance
as descending order. For each dataset, five used metrics are recorded in form of mean and standard deviation among different percentages. As \textbf{\textit{Water}} only have 16 features, we select 1 to 16 features. To show our performance more clearly, we also demonstrated a dataset for four metrics in Figure \ref{f3}. From the overall results,  following observations is obtained:

\begin{itemize}
    \item In terms of Marco-F1 and Mirco-F1, PML-FSLA ranks first on all eight datasets. As for Average Precision, Ranking Loss and Coverage, PML-FSLA ranks first except \textbf{\textit{CHD49}} and \textbf{\textit{Water}}. Among all the cases, PML-FSLA ranks first in 85\% of cases, and in only 7.5\% cases PML-FSLA doesn't rank in  top two. This result comprehensively proves the effectiveness of PML-FSLA.
    \item The superiority in Marco-F1 and Mirco-F1 proves the ability of PML-FSLA in dealing with sparsity and identifying positive labels. We attribute it to the reconstructed feature selection term. By matrix multiplication, the feature selection term \(QR\) represents the feature weight by the latent space similarity between the feature and the label, and eliminates the influence of unbalanced negative labels, so that the selected feature has strong identification ability for positive labels. The subsequent ablation experiments further confirmed this opinion.
    \item For the superiority in Average Precision, Ranking Loss and Coverage. We attribute it to the effective latent space alignment. Through this process, the information in the feature space and the the structural consistency between features and labels is utilized, and the noises in the label space is effectively eliminated, so the effect of the model is improved. As for the relatively poor effect on some datasets, it is because of the trade off of enhancing the positive label identification ability through feature selection. It leads to the weakening of the identification effect of negative labels, which is reflected in the decline of these metrics that show the overall effect.
\end{itemize}

\begin{figure}[htbp]
    \begin{subfigure}{0.32\linewidth}
        \centering
        \scalebox{1.3}{\includegraphics[width=\linewidth]{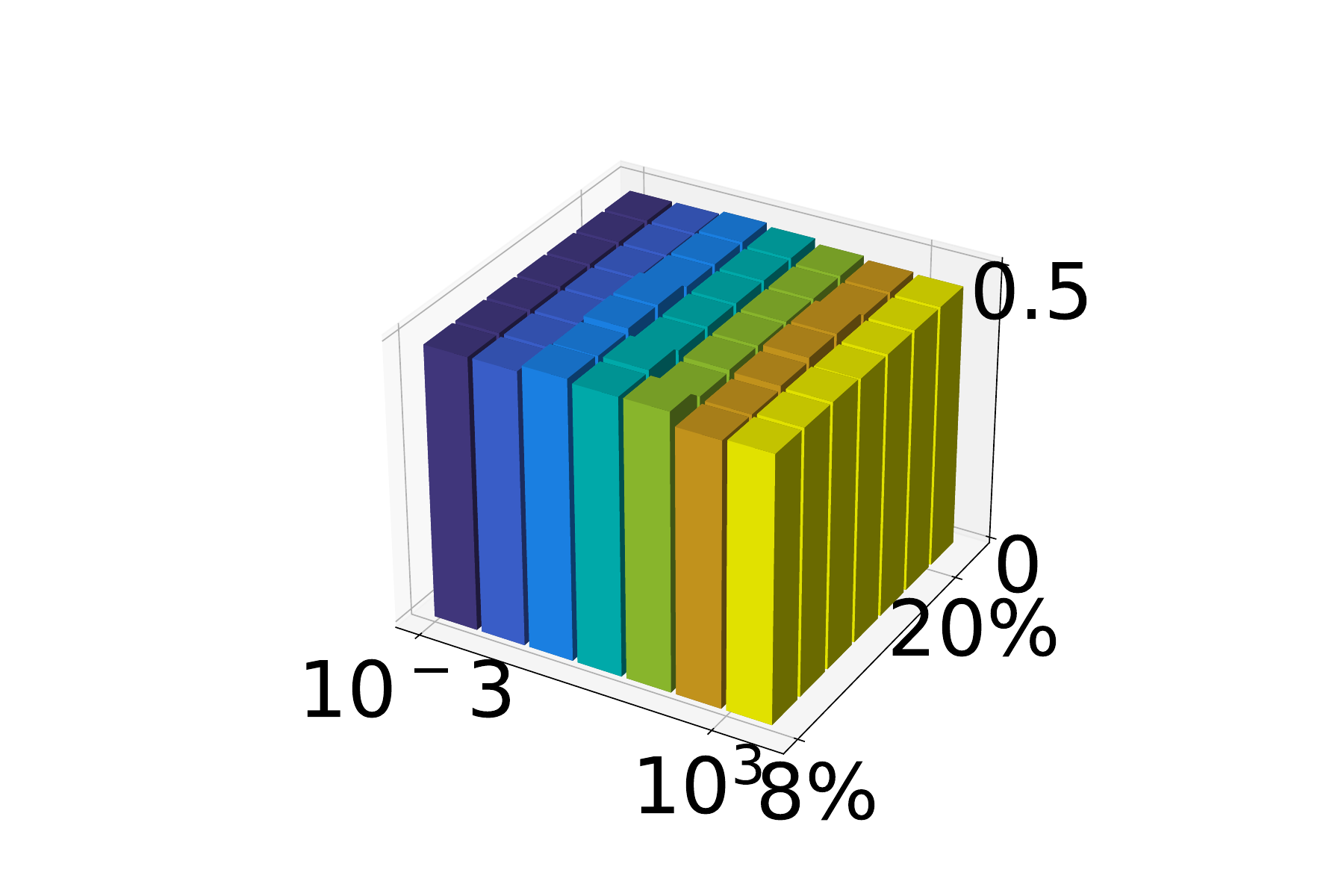}}
        \caption{$\alpha$}
        \label{psa}
    \end{subfigure}
    \centering
    \hspace{0.0001\linewidth}
    \begin{subfigure}{0.32\linewidth}
        \centering
        \scalebox{1.3}{\includegraphics[width=\linewidth]{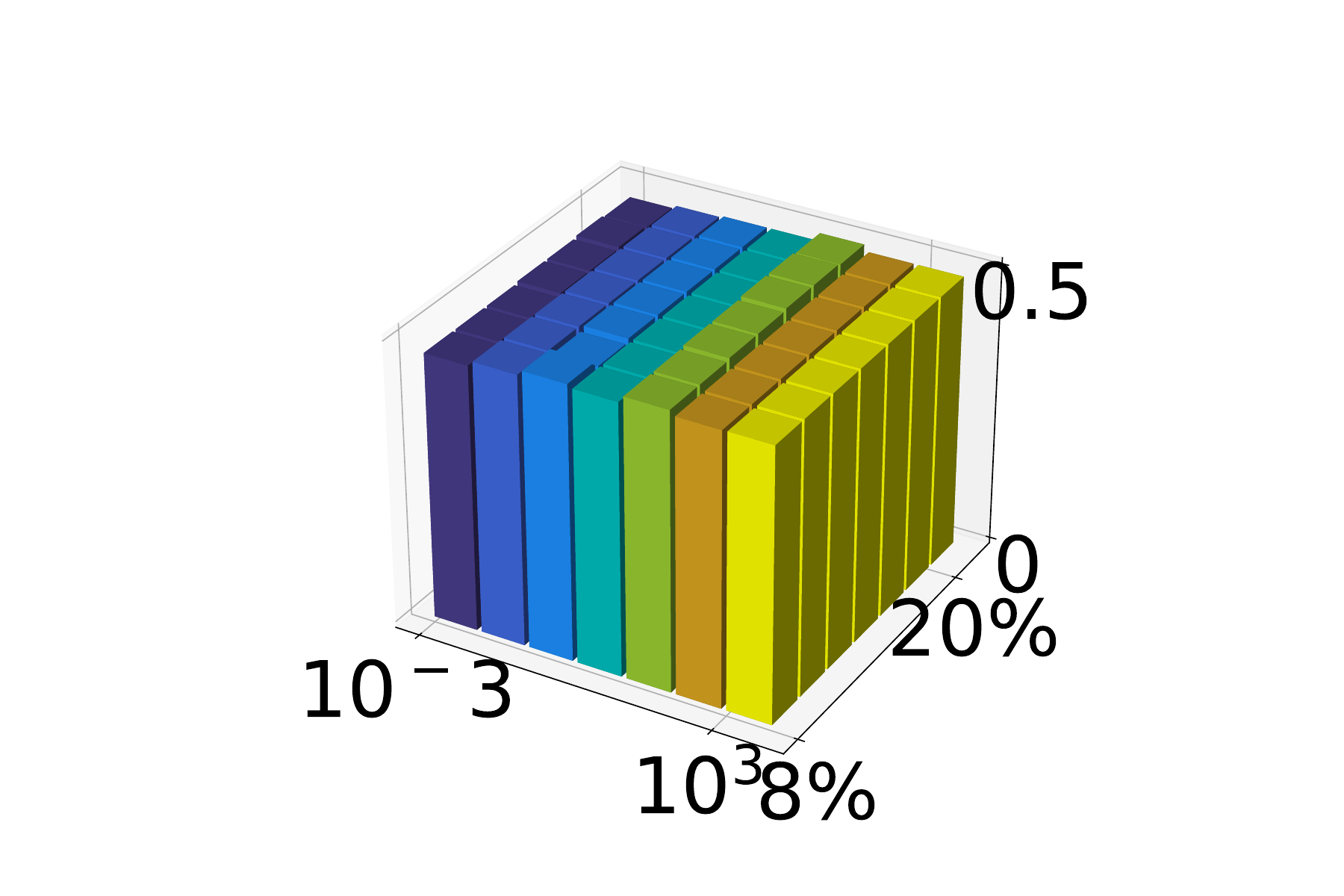}}
        \caption{$\beta$}
        \label{psb}
    \end{subfigure}
    \hspace{0.0001\linewidth}
    \begin{subfigure}{0.32\linewidth}
        \centering
        \scalebox{1.3}{\includegraphics[width=\linewidth]{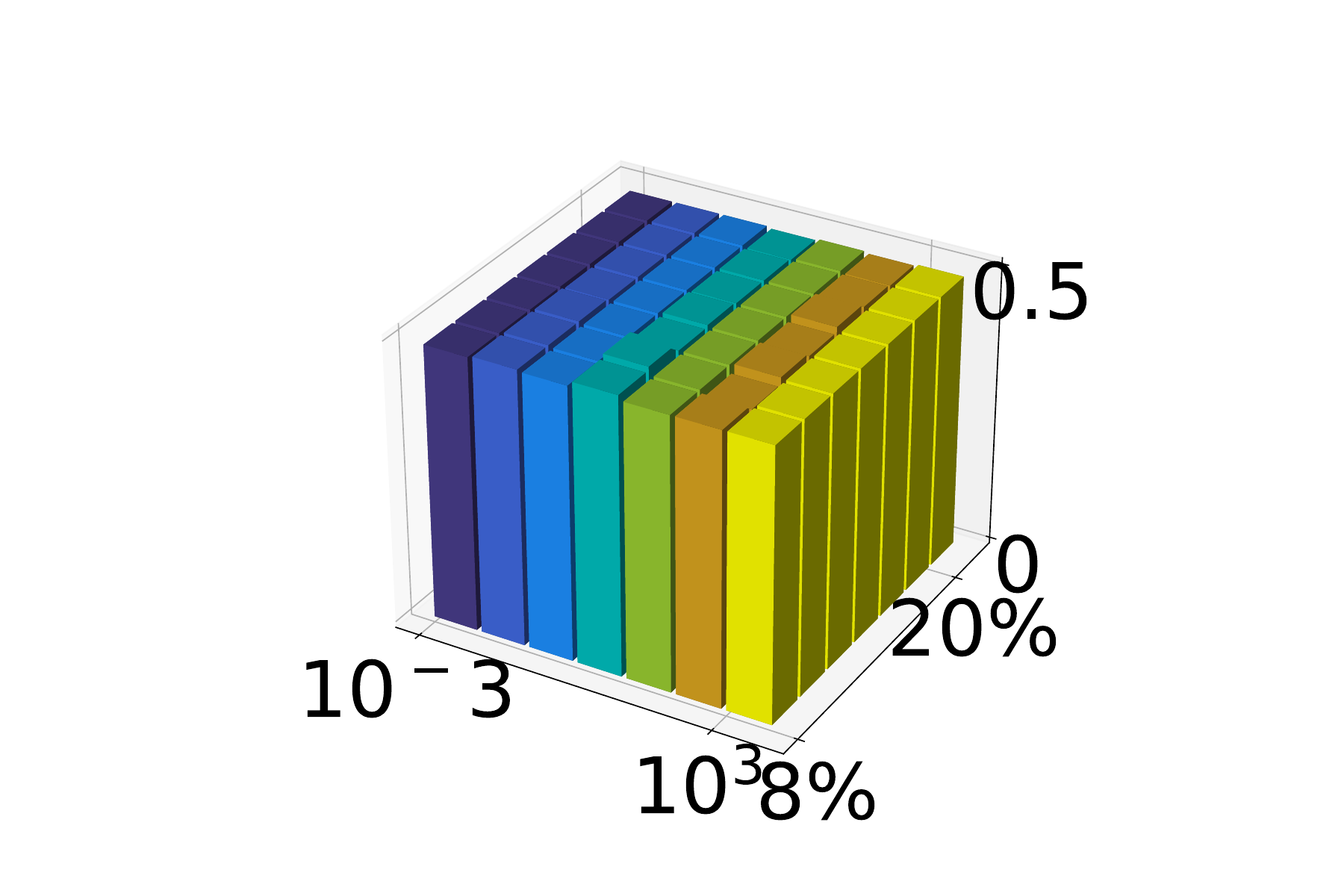}}
        \caption{$\gamma$}
        \label{psc}
    \end{subfigure}
    \caption{Parameter sensitivity studies on the \textbf{\textit{LLOG\_F}} in terms of Coverage.}
    \label{ps}
\end{figure}

\subsection{Parameter Analysis}
In PML-FSMIR, three parameters $\alpha$, $\beta$, and $\gamma$ could influence the experimental outcomes. Figure \ref{ps} illustrates the impact of these parameters on the model's performance on  \textbf{\textit{CHD\_49}} in terms of Ranking Loss. Each parameter was independently tuned over a range from 0.001 to 1000, and the model's performance was evaluated when selecting between 8\% to 20\% of the features. Ranking Loss decreases slightly as the number of selected features increases. It is evident that the model demonstrates a clear insensitivity to variations in these parameters, which highlights the robustness and stability of the model under different conditions.
\subsection{Ablation Study}
In order to prove that the feature selection term we designed is effective in enhancing the ability of the selected features to identify positive labels,  another feature selection term commonly used when involving latent space is employed as a comparison: the weight sparse matrix \(Q\) of the latent space projected to the feature space. Two methods are compared on seven datasets with Marco-F1 and Mirco-F1. The results are shown in Tables \ref{t7} and \ref{t8},  PML-FSLA-q is utilized to represent this comparison method utilizing the weight matrix \(Q\), and the larger value in each case is marked in bold.

In all cases, PML-FSLA beats PML-FSLA-q: the new feature selection term improves by at least 10\% over baseline. This result shows that only using the feature space projection weight matrix as feature selection term is insufficient. This method only considers the global structural consistency of the feature space and the label space, but cannot deal with the possible noises and local inconsistency problems. while both problems can be solved by comprehensively considering the projection weight matrices of the feature space and label space. Different from the above method that only relies on global structural consistency, this method measures the similarity of the feature space and label space through two weight matrices, so that when local inconsistency occurs, the corresponding weights in the feature selection terms will be adjusted to alleviate this problem. In addition, this method strongly strengthens the role of positive labels in the feature selection process because negative labels with a corresponding weight of zero will not be considered in this projection method. Although this operation will relatively reduce the effect of PML-FSLA in other evaluation metrics, the results demonstrate the trade off is worthwhile.
\begingroup
\setlength{\tabcolsep}{4pt}
\begin{table}[]
{\scriptsize
\begin{tabular}{@{}llllllll@{}}
\toprule
Datasets   & CAL            & CHD\_49        & Chess          & Corel5K        & LLOG F            & Water          & Yeast          \\ \midrule
PML-FSLA   & \textbf{0.538} & \textbf{0.637} & \textbf{0.579} & \textbf{0.263} & \textbf{0.339} & \textbf{0.556} & \textbf{0.496} \\
PML-FSLA-q & 0.149          & 0.477          & 0.000              & 0.000              & 0.022                 & 0.222          & 0.121          \\ \bottomrule
\end{tabular}
}
\caption{ Ablation experimental results of PML-FSLA in terms of Mirco-F1.}
\label{t7}
\end{table}

\begin{table}[]
{\scriptsize
\begin{tabular}{@{}llllllll@{}}
\toprule
Datasets   & CAL            & CHD\_49        & Chess          & Corel5K        & LLOG F            & Water          & Yeast          \\ \midrule
PML-FSLA   & \textbf{0.538} & \textbf{0.615} & \textbf{0.588} & \textbf{0.231} & \textbf{0.321} & \textbf{0.551} & \textbf{0.488} \\
PML-FSLA-q & 0.148          & 0.472          & 0.000              & 0.000              & 0.020                    & 0.212          & 0.132          \\ \bottomrule
\end{tabular}
}\caption{ Ablation experimental results of PML-FSLA in terms of Marco-F1.}
\label{t8}
\end{table}
\endgroup
\section{Conclusion}
In this paper,  the label disambiguation problem under the partial labeling scenario  is addressed within the latent space  by leveraging the inherent information  in  feature space as well as maintaining the structural consistency between the feature space and the label space. Additionally, rather than relying on traditional feature selection term commonly used in embedded feature selection methods when dealing with latent spaces, we start from another perspective and solve it by considering
the similarity of latent space projections. By doing so, we significantly enhance the model's capability to accurately identify positive labels, leading to more reliable and robust classification results. Extensive  experimental results demonstrate that our method outperforms existing approaches in various challenging scenarios, underscoring its effectiveness and robustness. 

\section*{Acknowledgments}
This work is funded by: by Science Foundation of Jilin Province of China under Grant No. 20230508179RC, and China Postdoctoral Science Foundation funded project under Grant No. 2023M731281 and Changchun Science and Technology Bureau Project 23YQ05.

\bibliography{aaai25}

\begin{thebibliography}{34}
\providecommand{\natexlab}[1]{#1}

\bibitem[{Abid, Khan, and Iqbal(2021)}]{abid2021review}
Abid, A.; Khan, M.~T.; and Iqbal, J. 2021.
\newblock A review on fault detection and diagnosis techniques: basics and beyond.
\newblock \emph{Artificial Intelligence Review}, 54(5): 3639--3664.

\bibitem[{Akbari and Hesamian(2019)}]{akbari2019elastic}
Akbari, M.~G.; and Hesamian, G. 2019.
\newblock Elastic net oriented to fuzzy semiparametric regression model with fuzzy explanatory variables and fuzzy responses.
\newblock \emph{IEEE Transactions on Fuzzy Systems}, 27(12): 2433--2442.

\bibitem[{Ankerst et~al.(1999)Ankerst, Breunig, Kriegel, and Sander}]{ankerst1999optics}
Ankerst, M.; Breunig, M.~M.; Kriegel, H.-P.; and Sander, J. 1999.
\newblock OPTICS: Ordering points to identify the clustering structure.
\newblock \emph{ACM Sigmod record}, 28(2): 49--60.

\bibitem[{Braytee et~al.(2017)Braytee, Liu, Catchpoole, and Kennedy}]{braytee2017multi}
Braytee, A.; Liu, W.; Catchpoole, D.~R.; and Kennedy, P.~J. 2017.
\newblock Multi-label feature selection using correlation information.
\newblock In \emph{Proceedings of the 2017 ACM on Conference on Information and Knowledge Management}, 1649--1656.

\bibitem[{Chen et~al.(2021)Chen, Li, Zeng, Zhang, Li, Huang, and Cai}]{chen2021predicting}
Chen, L.; Li, Z.; Zeng, T.; Zhang, Y.-H.; Li, H.; Huang, T.; and Cai, Y.-D. 2021.
\newblock Predicting gene phenotype by multi-label multi-class model based on essential functional features.
\newblock \emph{Molecular Genetics and Genomics}, 296(4): 905--918.

\bibitem[{Ester et~al.(1996)Ester, Kriegel, Sander, Xu et~al.}]{ester1996density}
Ester, M.; Kriegel, H.-P.; Sander, J.; Xu, X.; et~al. 1996.
\newblock A density-based algorithm for discovering clusters in large spatial databases with noise.
\newblock In \emph{kdd}, volume~96, 226--231.

\bibitem[{Gao, Li, and Hu(2021)}]{gao2021multilabel}
Gao, W.; Li, Y.; and Hu, L. 2021.
\newblock Multilabel feature selection with constrained latent structure shared term.
\newblock \emph{IEEE Transactions on Neural Networks and Learning Systems}, 34(3): 1253--1262.

\bibitem[{Hao, Hu, and Gao(2023)}]{hao2023partial}
Hao, P.; Hu, L.; and Gao, W. 2023.
\newblock Partial multi-label feature selection via subspace optimization.
\newblock \emph{Information Sciences}, 648: 119556.

\bibitem[{He et~al.(2019)He, Deng, Li, Shu, and Liu}]{he2019discriminatively}
He, S.; Deng, K.; Li, L.; Shu, S.; and Liu, L. 2019.
\newblock Discriminatively relabel for partial multi-label learning.
\newblock In \emph{2019 IEEE International Conference on Data Mining (ICDM)}, 280--288. IEEE.

\bibitem[{Hu et~al.(2020{\natexlab{a}})Hu, Li, Gao, and Zhang}]{2020Robust}
Hu, J.; Li, Y.; Gao, W.; and Zhang, P. 2020{\natexlab{a}}.
\newblock Robust multi-label feature selection with dual-graph regularization.
\newblock \emph{Knowledge-Based Systems}, 203: 106126.

\bibitem[{Hu et~al.(2020{\natexlab{b}})Hu, Li, Gao, Zhang, and Hu}]{hu2020multi}
Hu, L.; Li, Y.; Gao, W.; Zhang, P.; and Hu, J. 2020{\natexlab{b}}.
\newblock Multi-label feature selection with shared common mode.
\newblock \emph{Pattern Recognition}, 104: 107344.

\bibitem[{Jian et~al.(2016)Jian, Li, Shu, and Liu}]{2016Multi}
Jian, L.; Li, J.; Shu, K.; and Liu, H. 2016.
\newblock Multi-label informed feature selection.
\newblock In \emph{International Joint Conference on Artificial Intelligence}.

\bibitem[{Kumar et~al.(2023)Kumar, Koul, Singla, and Ijaz}]{kumar2023artificial}
Kumar, Y.; Koul, A.; Singla, R.; and Ijaz, M.~F. 2023.
\newblock Artificial intelligence in disease diagnosis: a systematic literature review, synthesizing framework and future research agenda.
\newblock \emph{Journal of ambient intelligence and humanized computing}, 14(7): 8459--8486.

\bibitem[{Li and Wang(2020)}]{li2020recovering}
Li, X.; and Wang, Y. 2020.
\newblock Recovering accurate labeling information from partially valid data for effective multi-label learning.
\newblock \emph{arXiv preprint arXiv:2006.11488}.

\bibitem[{Li, Lyu, and Feng(2021)}]{li2021partial}
Li, Z.; Lyu, G.; and Feng, S. 2021.
\newblock Partial multi-label learning via multi-subspace representation.
\newblock In \emph{Proceedings of the Twenty-Ninth International Conference on International Joint Conferences on Artificial Intelligence}, 2612--2618.

\bibitem[{Mokhtia, Eftekhari, and Saberi-Movahed(2021)}]{mokhtia2021dual}
Mokhtia, M.; Eftekhari, M.; and Saberi-Movahed, F. 2021.
\newblock Dual-manifold regularized regression models for feature selection based on hesitant fuzzy correlation.
\newblock \emph{Knowledge-Based Systems}, 229: 107308.

\bibitem[{Nie et~al.(2010)Nie, Huang, Cai, and Ding}]{nie2010efficient}
Nie, F.; Huang, H.; Cai, X.; and Ding, C. 2010.
\newblock Efficient and robust feature selection via joint l2, 1-norms minimization.
\newblock \emph{Advances in neural information processing systems}, 23.

\bibitem[{Pham et~al.(2022)Pham, Tao, Zhang, Yong, Li, and Xie}]{pham2022graph}
Pham, T.; Tao, X.; Zhang, J.; Yong, J.; Li, Y.; and Xie, H. 2022.
\newblock Graph-based multi-label disease prediction model learning from medical data and domain knowledge.
\newblock \emph{Knowledge-based systems}, 235: 107662.

\bibitem[{Qi et~al.(2018)Qi, Wang, Liu, Zhang, Wang, and Yi}]{qi2018unsupervised}
Qi, M.; Wang, T.; Liu, F.; Zhang, B.; Wang, J.; and Yi, Y. 2018.
\newblock Unsupervised feature selection by regularized matrix factorization.
\newblock \emph{Neurocomputing}, 273: 593--610.

\bibitem[{Shang et~al.(2020)Shang, Xu, Shang, and Jiao}]{shang2020sparse}
Shang, R.; Xu, K.; Shang, F.; and Jiao, L. 2020.
\newblock Sparse and low-redundant subspace learning-based dual-graph regularized robust feature selection.
\newblock \emph{Knowledge-Based Systems}, 187: 104830.

\bibitem[{Sun et~al.(2021)Sun, Feng, Liu, Lyu, and Lang}]{sun2021global}
Sun, L.; Feng, S.; Liu, J.; Lyu, G.; and Lang, C. 2021.
\newblock Global-local label correlation for partial multi-label learning.
\newblock \emph{IEEE Transactions on Multimedia}, 24: 581--593.

\bibitem[{Tibshirani(1996)}]{tibshirani1996regression}
Tibshirani, R. 1996.
\newblock Regression shrinkage and selection via the lasso.
\newblock \emph{Journal of the Royal Statistical Society Series B: Statistical Methodology}, 58(1): 267--288.

\bibitem[{Wei and Philip(2016)}]{wei2016unsupervised}
Wei, X.; and Philip, S.~Y. 2016.
\newblock Unsupervised feature selection by preserving stochastic neighbors.
\newblock In \emph{Artificial intelligence and statistics}, 995--1003. PMLR.

\bibitem[{Xie and Huang(2018)}]{xie2018partial}
Xie, M.-K.; and Huang, S.-J. 2018.
\newblock Partial multi-label learning.
\newblock In \emph{Proceedings of the AAAI conference on artificial intelligence}, volume~32.

\bibitem[{Xie and Huang(2020)}]{xie2020semi}
Xie, M.-K.; and Huang, S.-J. 2020.
\newblock Semi-supervised partial multi-label learning.
\newblock In \emph{2020 IEEE International Conference on Data Mining (ICDM)}, 691--700. IEEE.

\bibitem[{Xie and Huang(2021)}]{xie2021partial1}
Xie, M.-K.; and Huang, S.-J. 2021.
\newblock Partial multi-label learning with noisy label identification.
\newblock \emph{IEEE Transactions on Pattern Analysis and Machine Intelligence}, 44(7): 3676--3687.

\bibitem[{Xu, Liu, and Geng(2020)}]{xu2020partial}
Xu, N.; Liu, Y.-P.; and Geng, X. 2020.
\newblock Partial multi-label learning with label distribution.
\newblock In \emph{Proceedings of the AAAI conference on artificial intelligence}, volume~34, 6510--6517.

\bibitem[{Yilmaz et~al.(2021)Yilmaz, Kaynak, Ko{\c{c}}, Dibeklio{\u{g}}lu, and Kozat}]{yilmaz2021multi}
Yilmaz, S.~F.; Kaynak, E.~B.; Ko{\c{c}}, A.; Dibeklio{\u{g}}lu, H.; and Kozat, S.~S. 2021.
\newblock Multi-label sentiment analysis on 100 languages with dynamic weighting for label imbalance.
\newblock \emph{IEEE Transactions on Neural Networks and Learning Systems}.

\bibitem[{Yu et~al.(2018)Yu, Chen, Domeniconi, Wang, Li, Zhang, and Wu}]{yu2018feature}
Yu, G.; Chen, X.; Domeniconi, C.; Wang, J.; Li, Z.; Zhang, Z.; and Wu, X. 2018.
\newblock Feature-induced partial multi-label learning.
\newblock In \emph{2018 IEEE international conference on data mining (ICDM)}, 1398--1403. IEEE.

\bibitem[{Yu et~al.(2020)Yu, Yu, Wang, and Guo}]{yu2020partial}
Yu, T.; Yu, G.; Wang, J.; and Guo, M. 2020.
\newblock Partial multi-label learning with label and feature collaboration.
\newblock In \emph{Database Systems for Advanced Applications: 25th International Conference, DASFAA 2020, Jeju, South Korea, September 24--27, 2020, Proceedings, Part I 25}, 621--637. Springer.

\bibitem[{Zeng et~al.(2013)Zeng, Xiao, Jia, Chan, Gao, Xu, and Ma}]{zeng2013learning}
Zeng, Z.; Xiao, S.; Jia, K.; Chan, T.-H.; Gao, S.; Xu, D.; and Ma, Y. 2013.
\newblock Learning by associating ambiguously labeled images.
\newblock In \emph{Proceedings of the IEEE Conference on computer vision and pattern recognition}, 708--715.

\bibitem[{Zhang and Fang(2020)}]{zhang2020partial}
Zhang, M.-L.; and Fang, J.-P. 2020.
\newblock Partial multi-label learning via credible label elicitation.
\newblock \emph{IEEE Transactions on Pattern Analysis and Machine Intelligence}, 43(10): 3587--3599.

\bibitem[{Zhang and Zhou(2007)}]{2007ML}
Zhang, M.~L.; and Zhou, Z.~H. 2007.
\newblock ML-KNN: A lazy learning approach to multi-label learning.
\newblock \emph{Pattern Recognition}, 40(7): 2038--2048.

\bibitem[{Zhou et~al.(2018)Zhou, Zhang, Zhang, and He}]{zhou2018weakly}
Zhou, D.; Zhang, Z.; Zhang, M.-L.; and He, Y. 2018.
\newblock Weakly supervised POS tagging without disambiguation.
\newblock \emph{ACM Transactions on Asian and Low-Resource Language Information Processing (TALLIP)}, 17(4): 1--19.

\end{thebibliography}

\end{document}